\documentclass{article}

\usepackage{hyperref}  

\usepackage[preprint,nonatbib]{neurips_2025}

\usepackage[utf8]{inputenc}
\usepackage[T1]{fontenc}
\usepackage{hyperref}
\usepackage{url}
\usepackage{booktabs}
\usepackage{amsfonts}
\usepackage{nicefrac}
\usepackage{microtype}
\usepackage{xcolor}

\usepackage{mathtools}
\usepackage{amsmath}
\usepackage{amssymb}

\usepackage[longend]{algorithm2e}

\usepackage{graphicx}
\usepackage{caption}
\usepackage{subcaption}
\usepackage{tcolorbox}
\usepackage{float}
\usepackage{wrapfig}

\usepackage{tabularx}
\usepackage{tablefootnote}
\usepackage{array}

\usepackage{paralist}
\usepackage{appendix}
\usepackage{enumitem}

\usepackage{color}
\usepackage{soul}
\usepackage[htt]{hyphenat}
\usepackage{textgreek}

\usepackage{listings}

\definecolor{codegreen}{rgb}{0,0.6,0}
\definecolor{codegray}{rgb}{0.5,0.5,0.5}
\definecolor{codepurple}{rgb}{0.58,0,0.82}
\definecolor{codeyellow}{rgb}{0.67,0.67,0.0}
\definecolor{backcolour}{rgb}{0.95,0.95,0.92}

\lstdefinestyle{mystyle}{
    language=Python,
    backgroundcolor=\color{backcolour},   
    commentstyle=\color{codegreen},
    keywordstyle=\color{magenta},
    emph={testfunc,print,src},
    emphstyle=\color{codeyellow},
    numberstyle=\tiny\color{codegray},
    stringstyle=\color{codepurple},
    basicstyle=\footnotesize,
    breakatwhitespace=false,         
    breaklines=true,                 
    captionpos=b,                    
    keepspaces=true,                 
    numbers=left,                    
    numbersep=5pt,                  
    showspaces=false,                
    showstringspaces=false,
    showtabs=false,                  
    tabsize=2
}
\usepackage{doi}

\title{Oh That Looks Familiar: A Novel Similarity Measure for Spreadsheet Template Discovery}

\author{
  Anand Krishnakumar\thanks{Equal contribution} \\
  Ekimetrics \\
  \texttt{anand.krishnakumar@ekimetrics.com} \\
  \And
  Vengadesh Ravikumaran\footnotemark[1] \\
  Ekimetrics \\
  \texttt{vengadesh.ravikumaran@ekimetrics.com}
}

\begin{document}

\maketitle
\begin{abstract}
Traditional methods for identifying structurally similar spreadsheets fail to capture the spatial layouts and type patterns defining templates. To quantify spreadsheet similarity, we introduce a hybrid distance metric that combines semantic embeddings, data type information, and spatial positioning. In order to calculate spreadsheet similarity, our method converts spreadsheets into cell-level embeddings and then uses aggregation techniques like Chamfer and Hausdorff distances. Experiments across template families demonstrate superior unsupervised clustering performance compared to the graph-based Mondrian baseline, achieving perfect template reconstruction (Adjusted Rand Index of $1.00$ versus $0.90$) on the FUSTE dataset. Our approach facilitates large-scale automated template discovery, which in turn enables downstream applications such as retrieval-augmented generation over tabular collections, model training, and bulk data cleaning.
\end{abstract}


\section{Introduction}
Spreadsheets are ubiquitous in enterprises, yet collections are often messy and difficult to leverage at scale for LLM or ML applications. A key bottleneck is that similar spreadsheets—those following the same template or layout—are scattered across repositories, hindering automated processing and workflow integration.

We define spreadsheets as \emph{similar} if they share consistent header arrangements, data regions, and content distributions. Organizing spreadsheets by similarity enables enterprises to treat template families as unified objects—critical for emerging applications like table-based RAG systems, automated data wrangling pipelines, and foundation model pretraining over structured data.

Existing methods for spreadsheet similarity vary in their approaches: content-based embeddings \cite{copul2024tabee} focus primarily on semantic information while potentially overlooking layout structure, while graph-based approaches like Mondrian \cite{vitagliano2022mondrian} capture topological relationships through structural graphs. We propose a hybrid cell-level distance metric that jointly encodes spatial positioning, type patterns, and semantic content. By combining Euclidean layout similarity with type-aware semantic matching and aggregation strategies (Chamfer \cite{barrow1977parametric} and Hausdorff \cite{huttenlocher1993comparing} distances), our method effectively identifies spreadsheet template families for downstream processing.

The primary contribution of our paper is a \textbf{hybrid cell-level distance metric} for grouping spreadsheets into template families. We demonstrate superior unsupervised clustering performance compared to the graph-based Mondrian benchmark \cite{vitagliano2022mondrian}, achieving perfect template reconstruction.
\section{Related Work}
Our work intersects two primary research areas: spreadsheet representation methods and similarity measures. Prior spreadsheet understanding ranges from vision-based approaches~\cite{dong2019tablesense,deng2024tables} to sequential models~\cite{nishida2017understanding,gol2019tabular,wang2021tuta} and modern LLM encodings~\cite{zhang2023tablellama,li2023tablegpt,sui2023gpt4table}. Modern representation methods have explored various encoding formats (Markdown, HTML, JSON) for large language models~\cite{zhang2023tablellama, li2023tablegpt, sui2023gpt4table}, providing foundations for our encoding methodology, though they focus primarily on content rather than structural patterns. 

For similarity measurement, content-based methods~\cite{christodoulakis2020pytheas} focus on semantics while potentially underweighting spatial structure. Graph-based approaches like Mondrian~\cite{vitagliano2022mondrian}, our primary baseline, capture topology but exhibit a critical limitation: they consider content or structure independently. Our hybrid approach addresses this gap by jointly encoding spatial positioning, data types, and semantic content.
\section{Methodology}
\label{sec:methodology}
We measure spreadsheet similarity through hybrid distance metrics combining spatial layout, type information, and semantic content.

\subsection{Definitions}

\textbf{Definition 1 (Embedding).} 
For a spreadsheet $S$ with dimensions $m \times n$, we define its embedding as:
\begin{equation}
\label{def:phi}
    \Phi(S) = \{(i, j, t, s) : (i,j) \in [m] \times [n],\ t  \in \mathcal{T},\ s \in \mathbb{R}^{n}\}
\end{equation}
where $\mathcal{T}$ is a collection of data types $\{\text{Integer, Float, Date, String}, \ldots\}$.
Each element $(i, j, t, s) \in \mathbb{N}^2 \times \mathcal{T} \times \mathbb{R}^{n}$ represents a non-empty cell at position $(i,j)$ with data type $t$, where $s$ is a vector representation of the semantic meaning encoded using \textit{sentence-transformers/all-minilm-l6-v2} \cite{reimers2019sentence}. In our implementation, we map data types to integer encoding, details found in \ref{app:data_type_mapping}. This representation captures spatial positioning, type structure and semantic meaning.

\subsection{Cell-Level Distance (Hybrid Sub-Metric)}
\label{sec:cell_distance}

For two cells $u = (i_1, j_1, t_1,s_1)$ and $v = (i_2, j_2, t_2, s_2)$, we define:
\begin{equation}
\label{eq:cell_distance_matrix}
d_c =  w_\text{spatial} \cdot d_{\text{spatial}} + w_\text{type} \cdot d_{\text{type}} + w_\text{semantic} \cdot d_{\text{semantic}}
\end{equation}
where $w_\text{spatial},w_\text{semantic},w_\text{type} \in [0,1]$ and   $w_\text{spatial}+w_\text{semantic}+w_\text{type} = 1$. $d_c$ is a weighted average combination of the 3 dimensions.

\textbf{Spatial component:} Normalized Euclidean distance on cell positions

\begin{equation}
d_{\text{spatial}}(u,v) = \frac{\sqrt{(i_1 - i_2)^2 + (j_1 - j_2)^2}}{\sqrt{M_{\max}^2 + N_{\max}^2}}
\end{equation}

where $M_{\max}$ and $N_{\max}$ are the maximum row and column dimensions across both spreadsheets. This ensures $d_{\text{spatial}} \in [0,1]$ and makes distances comparable across different spreadsheet sizes.

\textbf{Type component:} Binary indicator for type mismatch
\begin{equation}
\label{eq: data type}
d_{\text{type}}(u,v) = \mathbf{1}_{t_1 \neq t_2} = \begin{cases} 0 & \text{if } t_1 = t_2 \\ 1 & \text{otherwise} \end{cases}
\end{equation}
\textbf{Semantic component:} Based on cosine similarity of cells
\begin{equation}
\label{eq:semantic_distance}
d_{\text{semantic}}(u,v) =  \frac{1}{2}\cdot (1 - \frac{s_1 \cdot s_2}{\|s_1\| \|s_2\|})
\end{equation}

Since all components are  normalized to $[0,1]$, their average $d_c$ is also a metric on $\mathbb{N}^2 \times \mathcal{T} \times \mathbb{R}^{n}$ with $d_c(u,v) \in [0,1]$.

\subsection{Spreadsheet-Level Distance (Aggregation Strategies)}

Given spreadsheets $S_1, S_2$ with embeddings $\Phi(S_1) = \{x_1, \ldots, x_m\}$ and $\Phi(S_2) = \{y_1, \ldots, y_n\}$, we aggregate cell-level distances $d_c$ to compute spreadsheet-level distances.Since $d_c$ is a metric, all aggregation methods inherit metric properties (symmetry, non-negativity, triangle inequality) on non-empty spreadsheets. We evaluate two aggregation strategies— Chamfer distance and Hausdorff distance which differ in their matching approaches. Formal definitions are provided in Appendix~\ref{app:aggregation}.

\section{Experiments}
\label{sec:experiments}

We evaluate our embedding framework and distance measure through three complementary sections: clustering analysis, relative importance of dimensions and computational scalability. Due to compute constraints, we restricted our analysis to 133 randomly selected spreadsheets across seven template families (catalog products, census, countries metadata, product manycols, and sport season, strategic focus, and triathlon) from the FUSTE real-world dataset \cite{vitagliano2022mondrian}, providing a reproducible benchmark for template discovery evaluation.
All code can be found here \url{https://anonymous.4open.science/r/spreadsheet-similarity-E286/README.md}.

\subsection{Clustering Analysis}
We evaluate the effectiveness of our structural embeddings for unsupervised organization of spread-
sheet collection by comparing our method with the Mondrian method proposed by  Vitagliano et al.\cite{vitagliano2022mondrian}. Using k-medoids clustering with k = 7 clusters (matching the number of template families in our dataset), we assess how well each method recovers the original template
structure.

\textbf{Results:} Table~\ref{tab:clustering_metrics} presents clustering performance across all distance measures. Our Chamfer-based method achieved perfect cluster recovery (ARI = $1.00$), substantially outperforming the Mondrian baseline (ARI = $0.90$) in partition quality. While Chamfer's silhouette coefficient ($0.64$) is lower than Mondrian's ($0.83$), the perfect ARI demonstrates superior recovery of the true cluster structure. Our Hausdorff-based approach underperformed both methods with ARI = $0.61$ and silhouette = $0.49$, suggesting the measure's sensitivity to outliers and extreme points makes it less suitable for this template discovery task. These results demonstrate that Chamfer distance combined with our hybrid similarity metric provides superior template discrimination. The perfect ARI confirms that incorporating semantic and type dimensions alongside spatial features enables the model to capture the essential structural characteristics that define document templates.

\begin{table}[h]
\centering
\caption{Clustering quality metrics using k-medoids with $k=7$ clusters. Initial values are $w_\text{semantic}=0.3$, $w_\text{type}=0.5$ and $w_\text{spatial}=0.2$. In \ref{subsec:rel_imp}, we vary them to study relative importance.}
\label{tab:clustering_metrics}
\begin{tabular}{lccc}
\toprule
\textbf{Distance Measure} & \textbf{ARI} (Adjusted Rand Index) $\uparrow$ &  \textbf{Silhouette Coeff.} \\
\midrule
\textbf{Chamfer} & \textbf{1.00} & 0.64 \\
\textbf{Mondrian} (benchmark) & 0.90 & 0.83 \\
\textbf{Hausdorff}& 0.61 & 0.49 \\
\bottomrule
\end{tabular}
\end{table}

\begin{figure}[h]
  \centering
  \includegraphics[width=\linewidth]{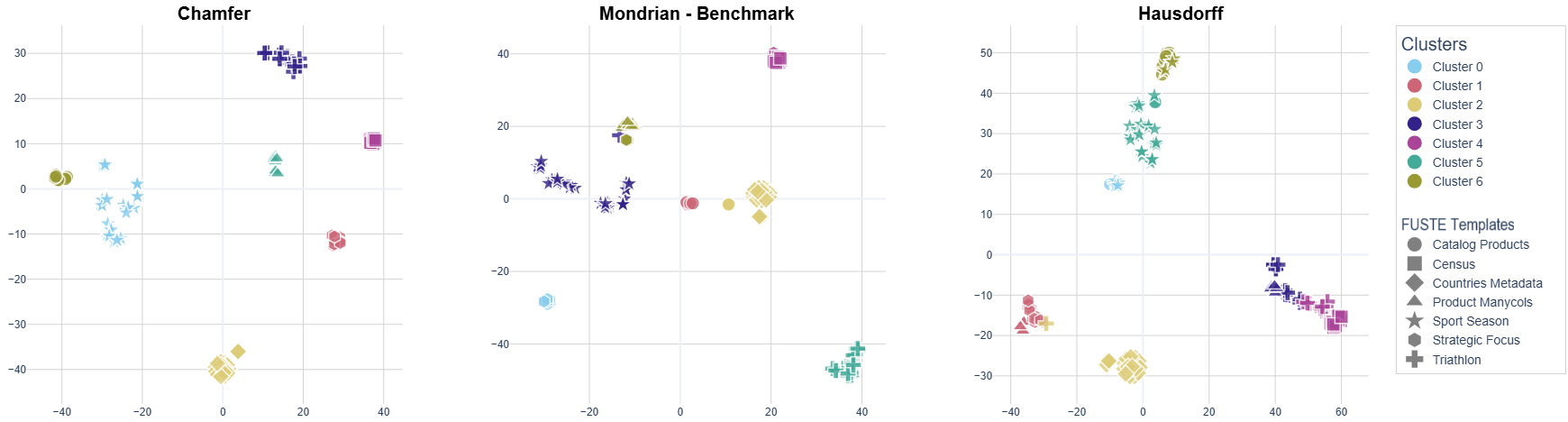}
  \caption{t-SNE projections (perplexity=$9$) of spreadsheet collections colored by k-medoids cluster assignments. Each subplot shows clustering results for a different distance measure, with point shapes indicating ground truth template families. Clear visual separation indicates successful cluster recovery.}
  \label{fig:umap_clusters}
\end{figure}

\subsection{Relative Importance of Dimensions}
\label{subsec:rel_imp}
To understand relative importance of spatial, type and semantic information in template discovery, we vary $w_\textit{type}, w_\textit{semantic}$ across the feasible domain and compute ARI and silhouette coefficients.

\begin{figure}[H]
\centering
\begin{subfigure}{0.5\textwidth}
  \centering
  \includegraphics[width=\linewidth]{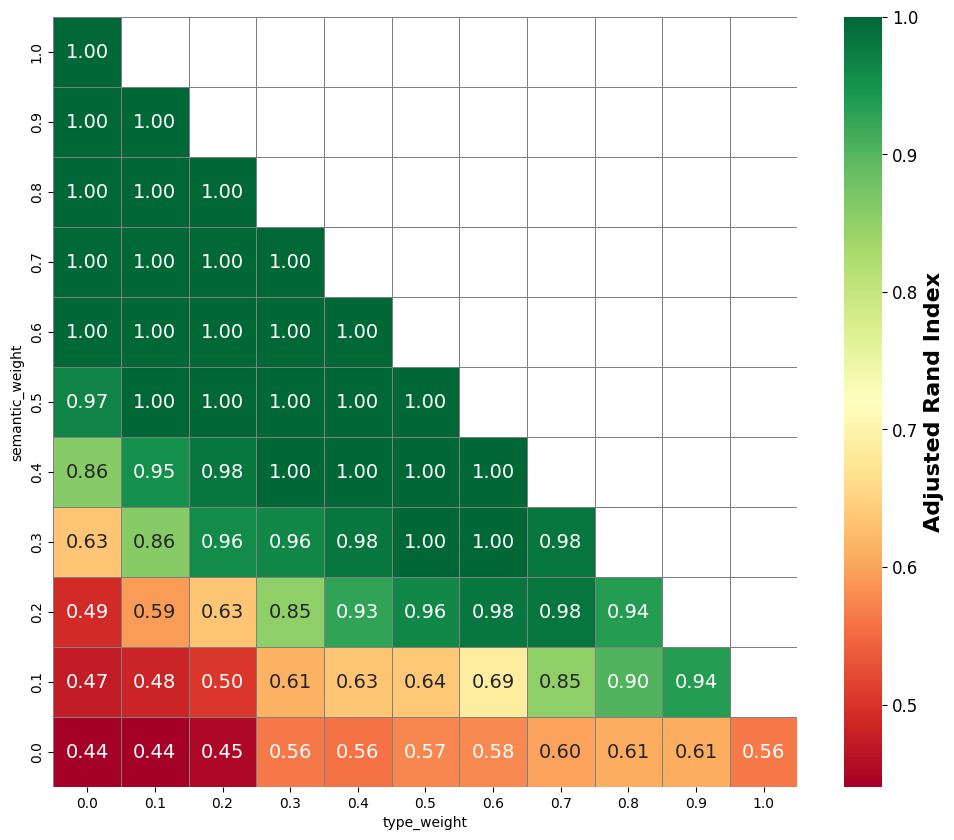}
  \caption{Adjusted Rand Index}
  \label{fig:ari_heatmap}
\end{subfigure}%
\begin{subfigure}{0.5\textwidth}
  \centering
  \includegraphics[width=\linewidth]{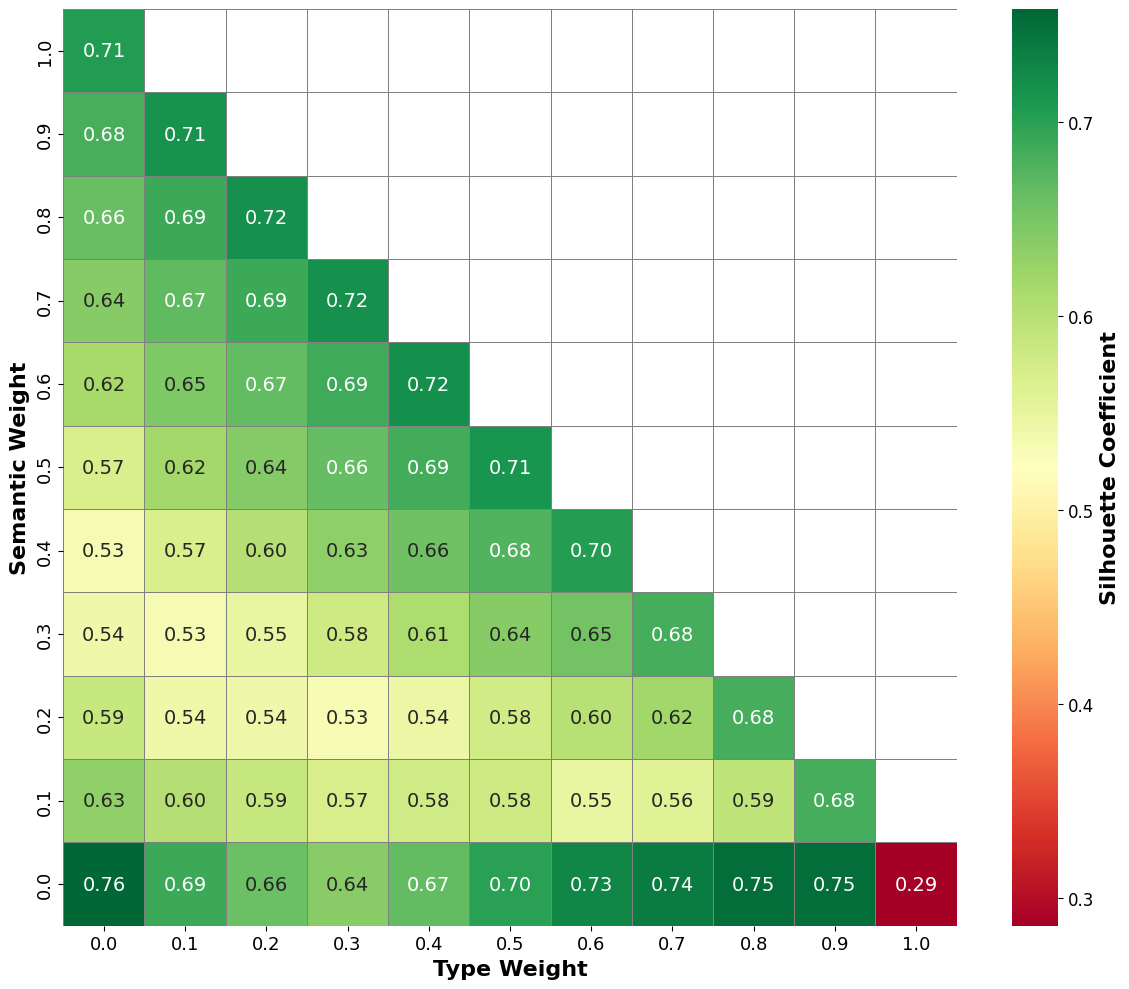}
  \caption{Silhouette Coefficient}
  \label{fig:silhouette_heatmap}
\end{subfigure}
\caption{Clustering performance across weight combinations for type and semantic dimensions. White cells indicate invalid combinations where $w_\textit{type} + w_\textit{semantic} > 1.0$.}
\label{fig:weight_heatmaps}
\end{figure}

The heatmaps in Figure~\ref{fig:weight_heatmaps} reveal key insights. First, \textbf{both dimensions contribute independently}: performance improves along both axes, with type weight enhancing ARI from $0.44$ to $0.61$ at $w_\textit{semantic}=0.0$, and semantic weight driving ARI from $0.44$ to $1.00$ at $w_\textit{type}=0.0$. However, \textbf{semantic information is most important}: the rate of improvement is much steeper along the semantic axis, with ARI jumping from $0.49$ to $0.85$ when $w_\textit{semantic}$ increases from $0.2$ to $0.3$ (at $w_\textit{type}=0.3$). Second, \textbf{the dimensions exhibit strong synergy at moderate weights}: type information amplifies the effect of semantic information (and vice versa). For instance, at $w_\textit{type}=0.2, w_\textit{semantic}=0.2$, ARI reaches $0.63$, substantially higher than pure type ($0.45$) or pure semantic ($0.49$) at weight $0.2$. The silhouette coefficient similarly shows complementary gains, increasing from $0.54$ to $0.68$ as type weight rises from $0.0$ to $0.4$ at $w_\textit{semantic}=0.3$. Third, \textbf{high semantic weight drives ARI performance}: once $w_\textit{semantic} \geq 0.5$, ARI reaches near-optimal levels (ARI $\geq 0.97$) regardless of type weight, demonstrating that semantic information alone is sufficient for accurate cluster recovery. However, \textbf{type weight enhances cluster cohesion}: the silhouette coefficient continues to improve with increasing type weight even at high semantic levels. For example, at $w_\textit{semantic}=0.7$, silhouette improves from $0.64$ (pure semantic-spatial) to $0.72$ (at $w_\textit{type}=0.3$), showing that type information contributes to tighter, more well-separated clusters. This reveals complementary roles: semantic information identifies the correct cluster assignments (external validity), while type information refines cluster quality and internal structure (internal validity).
\section{Conclusion}
\label{sec:conclusion}

This work presents a rigorous framework for quantifying similarity among spreadsheets through hybrid distance metrics that integrate spatial positioning with semantic-type information. Our primary contribution, \textbf{a novel hybrid distance metric}, showing superior clustering performance with ARI reaching $1.00$, surpassing benchmark methods.

\textbf{Limitations and Future Directions}: Our current framework establishes a foundation for embedding structural and semantic information from spreadsheets. We plan to improve scalability by optimising implementation, and to explore extending the approach to additional structured documents such as presentation slides.

\textbf{Broader Impact}: This work provides a principled methodology applicable to automated document organization and retrieval, template discovery, and data format standardization across extensive document collections.

Our framework establishes that visual structural patterns intuitively recognized by humans can be systematically quantified and leveraged to enhance performance in clustering and classification tasks, creating new opportunities for applications where classifying spreadsheets plays a vital role.

\begin{ack}
We thank Quentin Geoffroy for his technical assistance with experimental setup during his internship at Ekimetrics.
\end{ack}

\label{sec:references}

============================================================================
\appendix

\setcounter{table}{0}
\counterwithin{figure}{section}
\renewcommand{\thetable}{\Alph{section}\arabic{table}}

\section{Data Type Mapping}
\label{app:data_type_mapping}

For the structural embedding $\Phi(S)$ defined in Definition \ref{def:phi}, we map each cell's data type to an integer encoding to enable metric computation. Table~\ref{tab:type_mapping} presents the complete mapping used throughout our experiments.

\begin{table}[h]
\centering
\caption{Data type to integer encoding mapping.}
\label{tab:type_mapping}
\begin{tabular}{lc}
\toprule
Data Type & Encoding \\
\midrule
Integer & 0 \\
Float & 0 \\
Percentage & 0 \\
Scientific Notation & 0 \\
Currency & 0 \\
Date & 1 \\
Time & 1 \\
Email & 2 \\
Other & 3\\
String & 4 \\
\bottomrule
\end{tabular}
\end{table}

The type detection follows standard spreadsheet conventions: numerical formats are grouped together (e.g., scientific notation, currency symbols), temporal data by date/time patterns, and emails by the presence of the @ symbol. The ``Other'' category captures non-empty cells that do not match any specific type pattern. This encoding ensures that the type component $d_{\text{type}}(u,v)$ in \ref{eq: data type} operates on discrete categorical values while maintaining the metric structure required for our distance computations.
\section{Aggregation Strategies}
\label{app:aggregation}

Given spreadsheets $S_1, S_2$ with embeddings $\Phi(S_1) = \{x_1, \ldots, x_m\}$ and $\Phi(S_2) = \{y_1, \ldots, y_n\}$, we define 2 strategies for aggregating cell-level distances $d_c$ into spreadsheet-level distances:

\begin{itemize}
\item \textbf{Chamfer Distance:} Bidirectional average nearest-neighbor distance
$$D_{\text{Chamfer}}(S_1, S_2) = \frac{1}{m}\sum_{i=1}^m \min_{j} d_c(x_i, y_j) + \frac{1}{n}\sum_{j=1}^n \min_{j} d_c(x_i, y_j)$$

\item \textbf{Hausdorff Distance:} Worst-case nearest-neighbor distance
$$D_{\text{Hausdorff}}(S_1, S_2) = \max\left\{\max_i \min_j d_c(x_i, y_j), \max_j \min_i d_c(x_i, y_j)\right\}$$
\end{itemize}

\section{Theoretical Properties}

\noindent\textbf{Proposition 1 (Bounded Distance).} For any spreadsheets $S_1, S_2$ and aggregation method: $D(S_1, S_2) \in [0, 1]$.

\medskip
\noindent\textit{Proof.} Since $d_c \in [0,1]$, Chamfer, and Hausdorff are convex combinations of $d_c$ values.

\end{document}